\documentclass[11pt,a4paper]{article}
\usepackage[hyperref]{emnlp2020}
\usepackage{times}
\usepackage{latexsym}

\usepackage{amsmath}
\usepackage{amsfonts}
\usepackage{microtype}
\usepackage{graphicx}
\usepackage{multirow}
\usepackage{booktabs}
\usepackage{hhline}
\usepackage{xspace}
\usepackage{color}
\usepackage{enumitem}
\setlist[itemize]{leftmargin=*}
\setitemize[1]{itemsep=0pt,partopsep=0pt,parsep=\parskip,topsep=0pt}
\DeclareSymbolFont{extraup}{U}{zavm}{m}{n}
\DeclareMathSymbol{\varheart}{\mathalpha}{extraup}{86}
\DeclareMathSymbol{\vardiamond}{\mathalpha}{extraup}{87}

\aclfinalcopy

\title{Generating Radiology Reports via Memory-driven Transformer}

\author{
    Zhihong Chen$^{\spadesuit\heartsuit}$, \hspace{0.2cm}
    Yan Song$^{{\spadesuit}\heartsuit\dag}$, \hspace{0.2cm}
    Tsung-Hui Chang$^{\spadesuit\heartsuit}$, \hspace{0.2cm}
    Xiang Wan$^{\heartsuit}$ \\
    $^{\spadesuit}$The Chinese University of Hong Kong (Shenzhen)\\
    $^{\heartsuit}$Shenzhen Research Institute of Big Data\\
    $^{\spadesuit}$\texttt{zhihongchen@link.cuhk.edu.cn}\\
    $^{\spadesuit}$\texttt{\{songyan, changtsunghui\}@cuhk.edu.cn}\\
    $^{\heartsuit}$\texttt{wanxiang@sribd.cn}
}

\date{}

\begin{document}
\maketitle
\renewcommand{\thefootnote}{\fnsymbol{footnote}}
\footnotetext[2]{Corresponding author.}
\renewcommand{\thefootnote}{\arabic{footnote}}

\begin{abstract}
Medical imaging is frequently used in clinical practice and trials for diagnosis and treatment.
Writing imaging reports is time-consuming and can be error-prone for inexperienced radiologists.
Therefore, automatically generating radiology reports is highly desired to lighten the workload of radiologists and accordingly promote clinical automation, which
is an essential task to apply artificial intelligence to the medical domain.
In this paper, we propose to generate radiology reports with memory-driven Transformer, where a relational memory is designed to record key information of the generation process and a memory-driven conditional layer normalization is applied to incorporating the memory into the decoder of Transformer.
Experimental results on two prevailing radiology report datasets, IU X-Ray and MIMIC-CXR, show that our proposed approach outperforms previous models with respect to both language generation metrics and clinical evaluations.
Particularly, this is the first work reporting the generation results on MIMIC-CXR to the best of our knowledge.
Further analyses also demonstrate that our approach is able to generate long reports with necessary medical terms as well as meaningful image-text attention mappings.\footnote{Our code and the best performing models are released at \url{https://github.com/zhjohnchan/R2Gen}.}
\end{abstract}

\section{Introduction}
\label{sec:introduction}
Radiology report generation, which aims to automatically generate a free-text description for a clinical radiograph (e.g., chest X-ray), has emerged as a prominent attractive research direction in both artificial intelligence and clinical medicine.
It can greatly expedite the automation of workflows and improve the quality and standardization of health care. Recently, there are many methods proposed in this area \cite{coatt,hrgr,mimic,clinically,cmas}.

\begin{figure}[t]
\centering
\includegraphics[width=0.48\textwidth, trim=0 20 0 60]{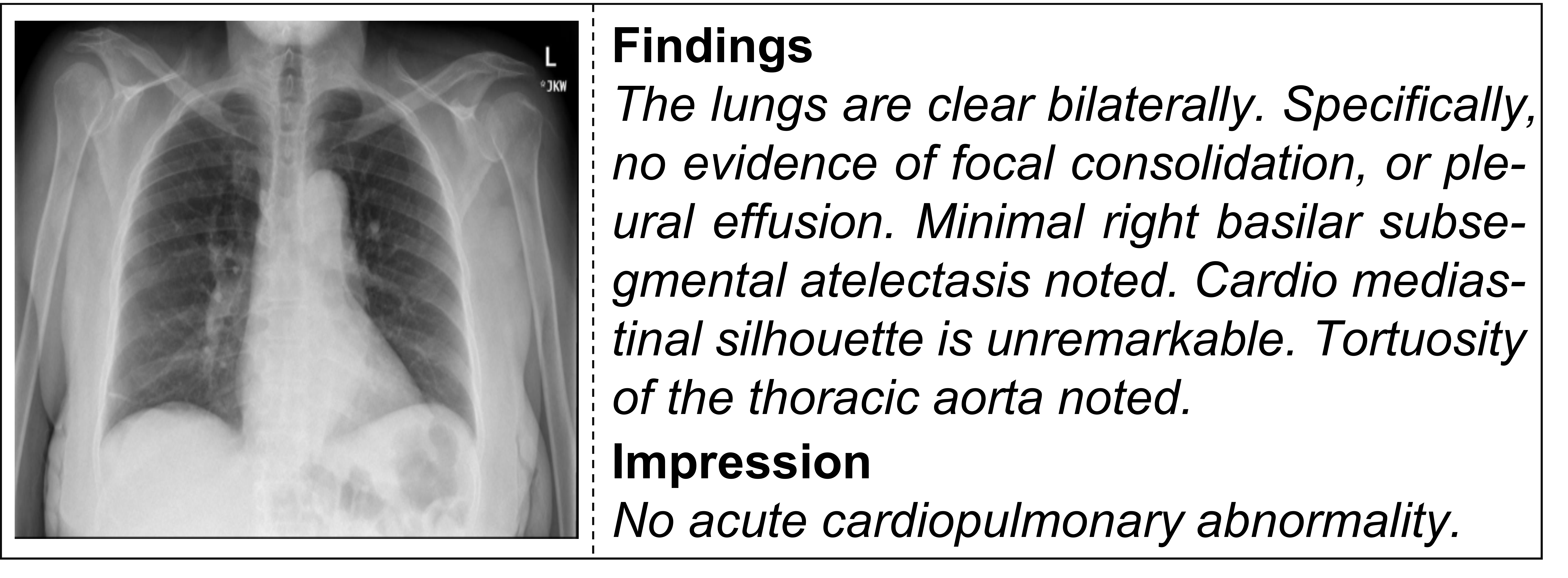}
\vspace{-0.6cm}
\caption{An example chest X-ray image and its report including findings and impression.}
\label{fig:example}
\vskip -1.5em
\end{figure}

Practically, a significant challenge of radiology report generation is that radiology reports are long narratives consisting of multiple sentences.
As illustrated by Figure \ref{fig:example}, a radiology report generally consists of a section of findings which describes medical observations, including both normal and abnormal features, as well as an impression or concluding remark summarizing the most prominent observations.
Therefore, applying conventional image captioning approaches \cite{showandtell,updown} may be insufficient for radiology report generation, as such approaches are designed to briefly describe visual scenes with short sentences.
The ability to provide accurate clinical descriptions for a radiograph is of the highest priority, which places a higher demand on the generation process.
Nevertheless, despite the difficulties posed by these evident length and accuracy requirements, radiology reports do have their own distinctive characteristics.
An important feature to note is their highly patternized nature, as illustrated by the sample report described above (Figure \ref{fig:example}).
On the basis of this patternization, many approaches have been proposed to address the challenges of radiology report generation.
For example, \newcite{clinically} found that a simple retrieval-based method could achieve a comparative performance for this task. \newcite{hrgr} combined retrieval-based and generation-based methods with manually extracted templates.
Although promising results may be obtained by the retrieval-based approaches, they are still limited in the preparation of large databases, or the explicit construction of template lists to determine the patterns embedded in various reports.

In this paper, we propose to generate radiology reports via memory-driven Transformer.
In detail,
a relational memory (RM) is proposed to record the information from previous generation processes and a novel memory-driven conditional layer normalization (MCLN) is designed to incorporate the relational memory into Transformer \cite{transformer}.
As a result,
similar patterns in different medical reports can be implicitly modeled and memorized during the generation process, which thereby can facilitate the decoding of Transformer and is capable of generating long reports with informative content.
Experimental results on two benchmark datasets confirm the validity and effectiveness of our approach, where Transformer with RM and MCLN achieves the state-of-the-art performance on all datasets.
To summarize, the contributions of this paper are four-fold:
\begin{itemize}
    \setlength{\topsep}{0pt}
    \setlength{\itemsep}{0pt}
    \setlength{\parsep}{0pt}
    \setlength{\parskip}{0pt}
    \item We propose to generate radiology reports via a novel memory-driven Transformer model.
    \item We propose a relational memory to record the previous generation process and the MCLN to incorporate relational memory into layers in the decoder of Transformer.
    \item Extensive experiments are performed and the results show that our proposed models outperform the baselines and existing models.
    \item We conduct analyses to investigate the effect of our model with respect to different memory sizes and show that our model is able to generate long reports with necessary medical terms and meaningful image-text attention mappings.
\end{itemize}

\section{The Proposed Method}
\label{sec:method}

\begin{figure*}[t]
\centering
\includegraphics[width=0.98\textwidth, trim=0 20 0 0]{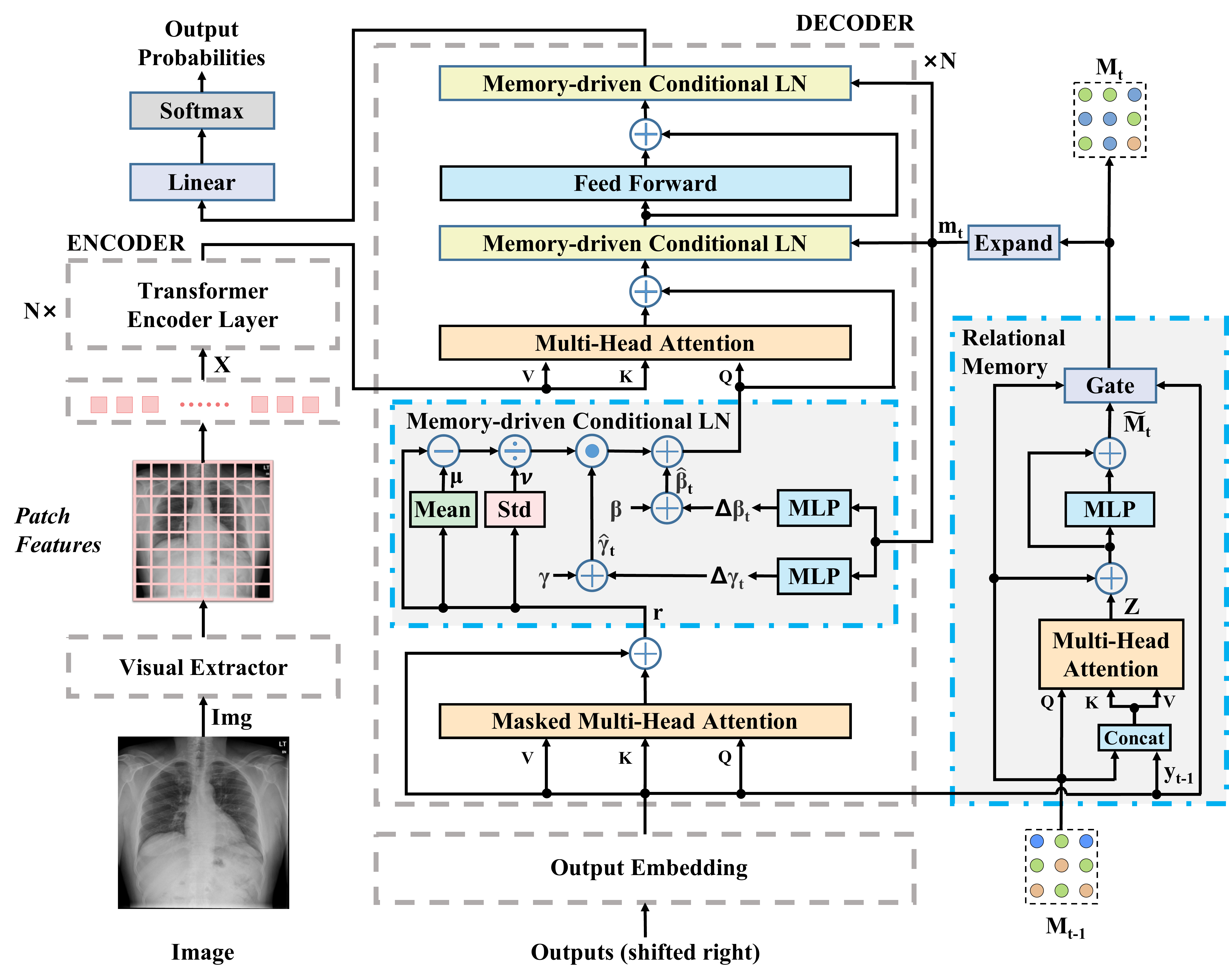}
\vspace{-0.2cm}
\caption{The overall architecture of our proposed model, where the visual extractor, encoder and decoder are shown in gray dash boxes and the details of the visual extractor and encoder are omitted.
The relational memory and memory conditional layer-normalization are illustrated in grey solid boxes with blue dash lines.}
\label{fig:model-architecture}
\vskip -1.05em
\end{figure*}

Generating radiology reports is essentially an image-to-text generation task,
for which there exist several solutions \cite{showandtell,xu2015show,updown,m2}.

We follow the standard sequence-to-sequence paradigm for this task.
In doing so, we treat the input from a radiology image as the source
sequence $\mathbf{X}=\{\mathbf{x}_{1}, \mathbf{x}_{2}, ..., \mathbf{x}_S\}, \mathbf{x}_{s}\in \mathbb{R}^{d}$, where $\mathbf{x}_{s}$ are patch features extracted from visual extractors and 
$d$ the size of the feature vector.
The corresponding report is the target sequence $Y=\{y_{1}, y_{2}, ..., y_{T}\}, y_{t}\in \mathbb{V}$, where $y_t$ are the generated tokens, $T$ the length of generated tokens and $\mathbb{V}$ the vocabulary of all possible tokens.
An overview of our proposed model is shown in Figure \ref{fig:model-architecture},
where the details are illustrated in following subsections.

\subsection{The Model Structure}
\label{subsec:preliminary}
Our model can be partitioned into three major components, i.e., the visual extractor, the encoder and the decoder, where the proposed memory and the integration of the memory into Transformer are mainly performed in the decoder.
The overall description of the three components and the training objective of the task is detailed below.

\vskip 0.5em

\noindent\textbf{Visual Extractor}~
Given a radiology image $Img$, its visual features $\mathbf{X}$ are extracted by pre-trained convolutional neural networks (CNN), e.g., VGG \cite{vgg} or ResNet \cite{resnet}, and the encoded results are used as the source sequence for all subsequent modules.
The process is formulated as:
\begin{equation}
\setlength\abovedisplayskip{6pt}
\setlength\belowdisplayskip{6pt}
    \{\mathbf{x}_{1}, \mathbf{x}_{2}, ..., \mathbf{x}_S\} = f_{v}(Img)
\end{equation}
where $f_{v}(\cdot)$ represents the visual extractor.
\vskip 0.5em

\noindent\textbf{Encoder}~
In our model, we use the standard encoder from Transformer, where the outputs are the hidden states $\mathbf{h}_{i}$ encoded from the input features $\mathbf{x}_{i}$ extracted from the visual extractor:
\begin{equation}
\setlength\abovedisplayskip{6pt}
\setlength\belowdisplayskip{6pt}
    \{\mathbf{h}_{1}, \mathbf{h}_{2}, ..., \mathbf{h}_{S}\} = f_{e}(\mathbf{x}_{1}, \mathbf{x}_{2}, ..., \mathbf{x}_S)
\end{equation}
where $f_{e}(\cdot)$ refers to the encoder.

\vskip 0.5em
\noindent\textbf{Decoder}~
The backbone decoder in our model is the one from Transformer, where we introduce an extra memory module to it by improving the original layer normalization with MCLN for each decoding layer as shown in Figure \ref{fig:model-architecture}.
Therefore the decoding process can be formalized as
\begin{equation}
    y_{t} = f_{d}(\mathbf{h}_{1}, ..., \mathbf{h}_{S}, \text{MCLN}(\text{RM}(y_{1}, ..., y_{t-1})))
\end{equation}
where $f_{d}(\cdot)$ refers to the decoder and the details of the memory (RM) and MCLN are presented in following subsections.

\vskip 0.5em
\noindent\textbf{Objective}~
Given the aforementioned structure, the entire generation process can be formalized as a recursive application of the chain rule
\begin{equation}
\setlength\abovedisplayskip{6pt}
\setlength\belowdisplayskip{6pt}
    p(Y|Img) = \prod_{t=1}^{T} p(y_{t}|y_{1}, ..., y_{t-1}, Img)
\end{equation}
where
$Y=\{y_{1}, y_{2}, ..., y_{T}\}$ is the target text sequence.
The model is then trained to maximize $P(Y|Img)$
through the negative conditional log-likelihood of $Y$ given the $Img$:
\begin{equation}
\setlength\abovedisplayskip{6pt}
\setlength\belowdisplayskip{6pt}
    \theta^{*} = \mathop{\arg\max}_{\theta} \sum_{t=1}^{T} \log p(y_{t}|y_{1}, ..., y_{t-1}, Img;\theta)
\end{equation}
where $\theta$ is the parameters of the model.

\vspace{-0.1cm}
\subsection{Relational Memory}
\label{subsec:relational-memory}
For any relevant $Img$, they may share similar patterns in their reports and they can be used as good references for each other to help the generation process.
As shown in Figure \ref{fig:example}, patterns such as ``\textit{The lungs are clear bilaterally}'' and ``\textit{no evidence of focal consolidation, or pleural effusion}'' always appear in the reports of similar images and are shown simultaneously.
To exploit such characteristics, we propose to use an extra component, i.e., relational memory, to enhance Transformer to
learn from the patterns and facilitate computing the interactions among patterns and the generation process.

In doing so, the relational memory uses a matrix to transfer its states over generation steps, where the states record important pattern information with each row (namely, memory slot) representing some pattern information.\footnote{Note that the rows (memory slots) and patterns do not follow one-to-one mapping, where the entire matrix serves as a whole unit to deliver the pattern information.}
During the generation,
the matrix is updated step-by-step with incorporating the output from previous steps.
Then, at time step $t$,
the matrix from the previous step, $\mathbf{M}_{t-1}$, is functionalized as the query and its concatenations with the previous output serve as the key and value to feed the multi-head attention module.
Given $H$ heads used in Transformer, there are $H$ sets of queries, keys and values via three linear transformations, respectively.
For each head, we obtain the query, key and value in the relational memory through $\mathbf{Q}=\mathbf{M}_{t-1}\cdot\mathbf{W_{q}}$, $\mathbf{K}=[\mathbf{M}_{t-1};\mathbf{y}_{t-1}]\cdot\mathbf{W_{k}}$ and $\mathbf{V}=[\mathbf{M}_{t-1};\mathbf{y}_{t-1}]\cdot\mathbf{W_{v}}$, respectively, where $\mathbf{y}_{t-1}$ is the embedding of the last output (at step $t-1$); $[\mathbf{M_{t-1}};\mathbf{y}_{t-1}]$ is the row-wise concatenation of $\mathbf{M}_{t-1}$ and $\mathbf{y}_{t-1}$. $\mathbf{W_{q}}$, $\mathbf{W_{k}}$ and $\mathbf{W_{v}}$ are the trainable weights of linear transformation of the query, key and value, respectively.
Multi-head attention is used to model $\mathbf{Q}$, $\mathbf{K}$ and $\mathbf{V}$
so as to depict relations of different patterns.
As a result,
\begin{equation}
\setlength\abovedisplayskip{6pt}
\setlength\belowdisplayskip{6pt}
    \mathbf{Z} = \text{softmax}(\mathbf{Q}\mathbf{K^{\top}}/\sqrt{d_{k}})\cdot\mathbf{V}
\end{equation}
where $d_{k}$ is the dimension of $\mathbf{K}$, and $\mathbf{Z}$ the output of the multi-head attention module.
Consider that the relational memory is performed in a recurrent manner along with the decoding process,
it potentially suffers from gradient vanishing and exploding.
We therefore introduce residual connections and a gate mechanism. The former is formulated as
\begin{equation}
    \mathbf{\Tilde{M}}_{t} = f_{mlp}(\mathbf{\mathbf{Z}} + \mathbf{M}_{t-1}) + \mathbf{Z} + \mathbf{M}_{t-1}
\end{equation}
where $f_{mlp}(\cdot)$ refers to the multi-layer perceptron (MLP).
\begin{figure}[t]
\centering
\includegraphics[width=0.48\textwidth, trim=0 30 0 30]{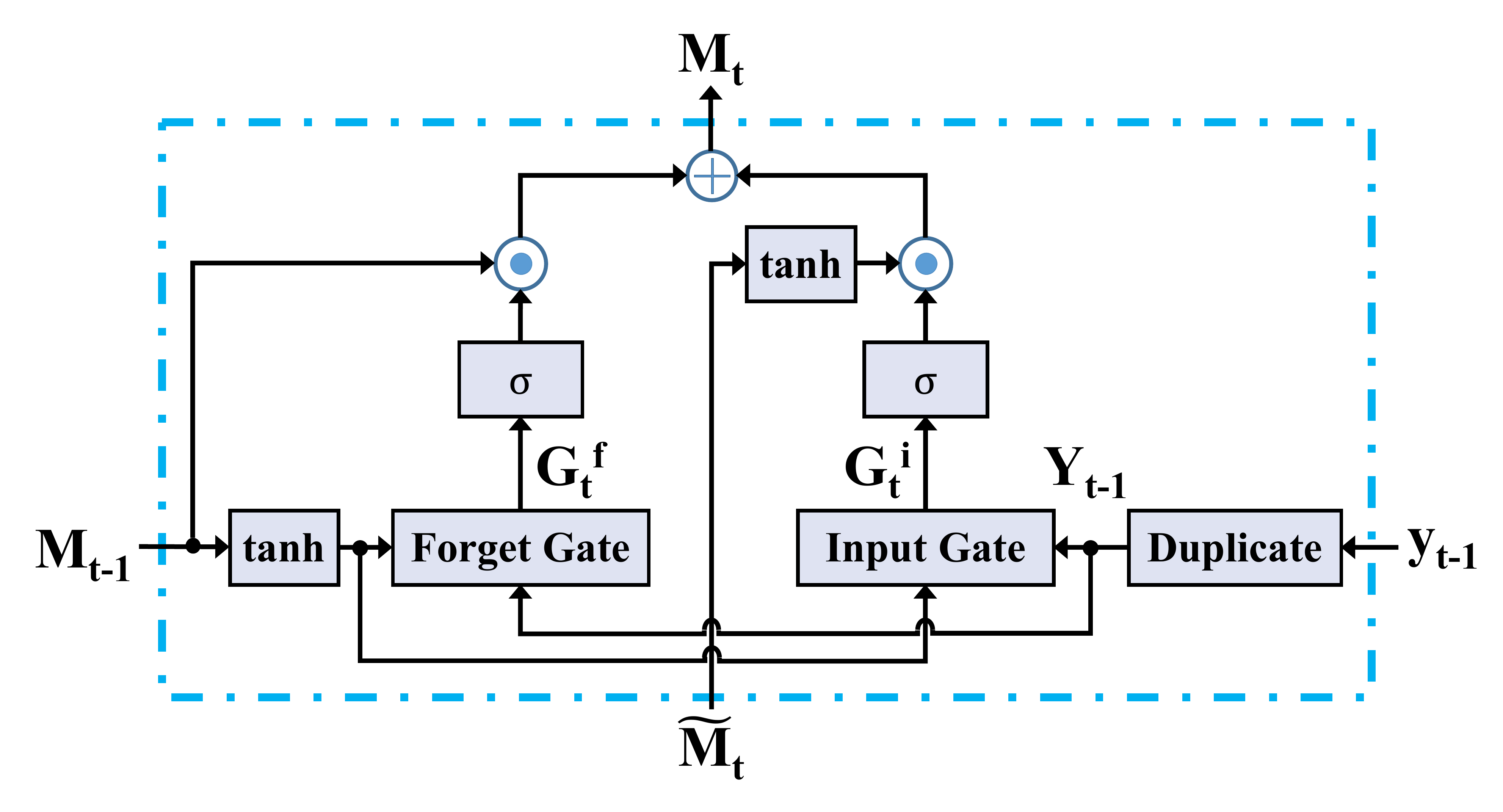}
\vskip -0.5em
\caption{The illustration of the gate mechanism.}
\label{fig:gated-operation}
\vskip -1em
\end{figure}
The detailed structure of the gate mechanism in the relational memory is shown in Figure \ref{fig:gated-operation},
where the forget and input gates are applied to balance the inputs from $\mathbf{M}_{t-1}$ and $\mathbf{y}_{t-1}$, respectively.
To ensure that $\mathbf{y}_{t-1}$ can be used for computation with $\mathbf{M}_{t-1}$,
it is extended to a matrix $\mathbf{Y}_{t-1}$ by duplicating it to multiple rows.
Therefore, the forget and input gate are formalized as
\begin{align}
\mathbf{G}_{t}^{f} &= \mathbf{Y}_{t-1} \mathbf{W}^{f} + \text{tanh}(\mathbf{M}_{t-1})\cdot\mathbf{U}^{f} \\
\mathbf{G}_{t}^{i} &= \mathbf{Y}_{t-1} \mathbf{W}^{i} + \text{tanh}(\mathbf{M}_{t-1})\cdot\mathbf{U}^{i}
\end{align}
where $\mathbf{W}^{f}$ and $\mathbf{W}^{i}$ are trainable weights for $\mathbf{Y}_{t-1}$ in each gate; 
similarly, $\mathbf{U}^{f}$ and $\mathbf{U}^{i}$ are the trainable weights for $\mathbf{M}_{t-1}$ in each gate.
The final output of the gate mechanism is formalized as
\begin{align}
    \mathbf{M}_{t} = \sigma(\mathbf{G}_{t}^{f}) \odot \mathbf{M}_{t-1} + \sigma(\mathbf{G}_{t}^{i}) \odot \text{tanh}(\mathbf{\Tilde{M}}_{t})
\end{align}
where $\odot$ refers to the Hadamard product and $\sigma$ the sigmoid function and $\mathbf{M}_{t}$ is the output of the entire relational memory module at step $t$.

\begin{table}[t]
\footnotesize
\centering
\setlength{\tabcolsep}{1mm}{\begin{tabular}{@{}l|rrr|rrr@{}}
\toprule
\multirow{2}{*}{\textsc{\textbf{Dataset}}} & \multicolumn{3}{c|}{\textsc{\textbf{IU X-Ray}}}                                                                     & \multicolumn{3}{c}{\textsc{\textbf{MIMIC-CXR}}}                                                                    \\ \cmidrule(l){2-7} 
                                  & \multicolumn{1}{c}{\textsc{Train}} & \multicolumn{1}{c}{\textsc{Val}} & \multicolumn{1}{c|}{\textsc{Test}} & \multicolumn{1}{c}{\textsc{Train}} & \multicolumn{1}{c}{\textsc{Val}} & \multicolumn{1}{c}{\textsc{Test}} \\ \midrule
\textsc{Image \#}                 & 5,226                              & 748                              & 1,496                                & 368,960                            & 2,991                            & 5,159                             \\
\textsc{Report \#}                & 2,770                              & 395                              & 790                                & 222,758                            & 1,808                            & 3,269                             \\
\textsc{Patient \#}               & 2,770                              & 395                              & 790                                & 64,586                             & 500                              & 293                               \\
\textsc{Avg. Len.}                & 37.56                              & 36.78                            & 33.62                              & 53.00                              & 53.05                            & 66.40                             \\ \bottomrule
\end{tabular}}
\vskip -0.25em
\caption{The statistics of the two benchmark datasets w.r.t. their training, validation and test sets, including the numbers of images, reports and patients, and the average word-based length (\textsc{Avg. Len.}) of reports. }
\label{table:datasets}
\vskip -1em
\end{table}
\begin{table*}[t]
\centering
\setlength{\tabcolsep}{1.33mm}{\begin{tabular}{@{}l|l|ccccccc|ccc@{}}
\toprule
\multirow{2}{*}{\textsc{\textbf{Data}}}                                             & \multicolumn{1}{c|}{\multirow{2}{*}{\textsc{\textbf{Model}}}} & \multicolumn{7}{c|}{\textsc{\textbf{NLG Metrics}}}                                                              & \multicolumn{3}{c}{\textsc{\textbf{CE Metrics}}} \\
                                                                                       & \multicolumn{1}{c|}{}                                         & \textsc{BL-1}  & \textsc{BL-2}  & \textsc{BL-3}  & \textsc{BL-4}  & \textsc{MTR}   & \textsc{RG-L}  & \textsc{Avg. $\Delta$} & \textsc{P}        & \textsc{R}       & \textsc{F1}      \\ \midrule
\multirow{3}{*}{\begin{tabular}[c]{@{}l@{}}\textsc{IU}\\ \textsc{X-Ray}\end{tabular}}  & \textsc{Base}                                              & 0.396          & 0.254          & 0.179          & 0.135          & 0.164          & 0.342          & -                      & -                 & -                & -                \\
                                                                                       & \textsc{~+rm}                                               & 0.444          & 0.283          & 0.196          & 0.141          & 0.179          & 0.364          & \ \ 8.9\%              & -                 & -                & -                \\
                                                                                       & \textsc{~+rm+mcln}                                                 & \textbf{0.470} & \textbf{0.304} & \textbf{0.219} & \textbf{0.165} & \textbf{0.187} & \textbf{0.371} & \textbf{17.6\%}        & -                 & -                & -                \\ \midrule
\multirow{3}{*}{\begin{tabular}[c]{@{}l@{}}\textsc{MIMIC}\\ \textsc{-CXR}\end{tabular}} & \textsc{Base}                                              & 0.314          & 0.192          & 0.127          & 0.090          & 0.125          & 0.265          & -                      & 0.331             & 0.224            & 0.228            \\
                                                                                       & \textsc{~+rm}                                               & 0.330          & 0.200          & 0.133          & 0.095          & 0.128          & 0.265          & \ \ 3.7\%              & 0.325             & 0.243            & 0.249            \\
                                                                                       & \textsc{~+rm+mcln}                                                 & \textbf{0.353} & \textbf{0.218} & \textbf{0.145} & \textbf{0.103} & \textbf{0.142} & \textbf{0.277} & \textbf{12.1\%}        & \textbf{0.333}    & \textbf{0.273}   & \textbf{0.276}   \\ \bottomrule
\end{tabular}}
\vskip -0.3em
\caption{The performance of all baselines and our full model on the test sets of \textsc{IU X-Ray} and \textsc{MIMIC-CXR} datasets with respect to NLG and CE metrics.
BL-n denotes BLEU score using up to n-grams; MTR and RG-L denote METEOR and ROUGE-L, respectively.
The average improvement over all NLG metrics compared to \textsc{Base} is also presented in the ``\textsc{Avg. $\Delta$}'' column.
The performance of all models is averaged from five runs.}
\label{table:baselines}
\vskip -1.05em
\end{table*}

\subsection{Memory-driven Conditional Layer Normalization}
\label{subset:mcln}
Although memory shows its effectiveness in many NLP tasks \cite{end2end,productkeys}, it is by default applied to encoding with rather isolated designs.
However, given that text generation is a dynamic process and largely affected by the output at each decoding step, memory is expected to be closely integrated to the decoder.

Therefore,
we propose a novel
MCLN and use it to incorporate the relational memory to enhance the decoding of Transformer.
Recall that in the conventional Transformer, to improve generalization, $\gamma$ and $\beta$ are two crucial parameters for scaling and shifting the learned representations,\footnote{In detail, $\gamma$ is used to amplify the values in the learned representation and $\beta$ provides a bias adjustment to them.} respectively.
Thus we propose to incorporate the relational memory via MCLN by feeding its output $\mathbf{M}_t$ to $\mathbf{\gamma}$ and $\mathbf{\beta}$.
Consequently, this design takes the benefit from the memory while preventing it from influencing too many parameters of Transformer so that some core information for generation is not affected.

As shown in Figure \ref{fig:model-architecture},
in each Transformer decoding layer, we use three MCLNs,
where the output of the first MCLN is functionalized as the query to be fed into the following multi-head attention module together with the hidden states from the encoder as the key and value.
To feed each MCLN, at step $t$,
the output of the relational memory $\mathbf{M}_t$ is expanded into a vector $\mathbf{m}_{t}$ by simply concatenating all rows from $\mathbf{M}_t$.
Then, an MLP is used to predict a change $\Delta \mathbf{\gamma}_{t}$ on $\mathbf{\gamma}_{t}$ from $\mathbf{m}_{t}$, and update it via
\begin{align}
    \Delta \mathbf{\gamma}_{t} &= f_{mlp}(\mathbf{m}_{t})\\
    \mathbf{\hat{\gamma}}_{t} &= \mathbf{\gamma} + \Delta \mathbf{\gamma}_{t}
\end{align}
Similarly, $\Delta \mathbf{\beta}_{t}$ and $\mathbf{\hat{\beta}}_{t}$ are performed by
\begin{align}
    \Delta \mathbf{\beta}_{t} &= f_{mlp}(\mathbf{m}_{t})\\
    \mathbf{\hat{\beta}}_{t} &= \mathbf{\beta} + \Delta \mathbf{\beta}_{t}
\end{align}
Afterwards, the predicted $\mathbf{\hat{\beta}}_{t}$ and $\mathbf{\hat{\gamma}}_{t}$ are applied to the
mean and variance results
of the multi-head self-attention from the previous generated outputs:
\begin{equation}
    f_{mcln}(\mathbf{r}) = \mathbf{\hat{\gamma}}_{t} \odot  \frac{\mathbf{r}-\mu}{\upsilon} +  \mathbf{\hat{\beta}}_{t}
\label{formula:mcln}
\end{equation}
where $\mathbf{r}$ refers to the output
from the previous module;
$\mu$ and $\upsilon$ are the mean and standard deviation of $\mathbf{r}$, respectively.
The result $f_{mcln}(\mathbf{r})$ from MCLN is then fed to the next module (for the 1st and 2nd MCLN) or used as the final output for generation (for the 3rd MCLN).

\section{Experiment Settings}
\label{sec:experiments}
\subsection{Datasets}
\label{subsec:datasets}
We conduct our experiments on two datasets, which are described as follows:
\begin{itemize}
    \setlength{\topsep}{0pt}
    \setlength{\itemsep}{0pt}
    \setlength{\parsep}{0pt}
    \setlength{\parskip}{0pt}
    \item \textbf{\textsc{IU X-Ray}} \cite{iuxray}\footnote{\url{https://openi.nlm.nih.gov/}}: a public radiography dataset collected by Indiana University with 7,470 chest X-ray images and 3,955 reports.
    \item \textbf{\textsc{MIMIC-CXR}} \cite{mimic}\footnote{\url{https://physionet.org/content/mimic-cxr/2.0.0/}}: the largest radiology dataset to date that consists of 473,057 chest X-ray images and 206,563 reports from 63,478 patients.
\end{itemize}
For both datasets, we follow \newcite{hrgr} to exclude the samples without reports. Then we apply their conventional splits.
Specifically, \textsc{IU X-Ray} is partitioned into train/validation/test set by 
7:1:2 of the entire dataset,
and \textsc{MIMIC-CXR}'s official split
is adopted.
The statistics of the datasets are shown in Table \ref{table:datasets}, with the numbers of images, reports, patients and the average length of reports.

\begin{table*}[t]
\centering
\begin{tabular}{@{}l|l|cccccc|ccc@{}}
\toprule
\multirow{2}{*}{\textsc{\textbf{Data}}}                                             & \multicolumn{1}{c|}{\multirow{2}{*}{\textsc{\textbf{Model}}}} & \multicolumn{6}{c|}{\textsc{\textbf{NLG Metrics}}}                                     & \multicolumn{3}{c}{\textsc{\textbf{CE Metrics}}} \\
                                                                                       & \multicolumn{1}{c|}{}                                         & \textsc{BL-1}  & \textsc{BL-2}  & \textsc{BL-3}  & \textsc{BL-4}  & \textsc{MTR}   & \textsc{RG-L}  & \textsc{P}     & \textsc{R}    & \textsc{F1}      \\ \midrule
\multirow{7}{*}{\begin{tabular}[c]{@{}l@{}}\textsc{IU}\\ \textsc{X-Ray}\end{tabular}}  & \textsc{ST}$^{\natural}$                                      & 0.216          & 0.124          & 0.087          & 0.066          & -              & 0.306          & -                 & -                & -                \\
                                                                                       & \textsc{Att2in}$^{\natural}$                                  & 0.224          & 0.129          & 0.089          & 0.068          & -              & 0.308          & -                 & -                & -                \\
                                                                                       & \textsc{AdaAtt}$^{\natural}$                                  & 0.220          & 0.127          & 0.089          & 0.068          & -              & 0.308          & -                 & -                & -                \\ \cmidrule(l){2-11} 
                                                                                       & \textsc{CoAtt}$^{\natural}$                                   & 0.455          & 0.288          & 0.205          & 0.154          & -              & 0.369          & -                 & -                & -                \\
                                                                                       & \textsc{Hrgr}$^{\natural}$                                    & 0.438          & 0.298          & 0.208          & 0.151          & -              & 0.322          & -                 & -                & -                \\
                                                                                       & \textsc{Cmas-RL}$^{\natural}$                                 & 0.464          & 0.301          & 0.210          & 0.154          & -              & 0.362          & -                 & -                & -                \\ \cmidrule(l){2-11} 
                                                                                       & \textsc{Ours}                                                 & \textbf{0.470} & \textbf{0.304} & \textbf{0.219} & \textbf{0.165} & 0.187          & \textbf{0.371} & -                 & -                & -                \\ \midrule
\multirow{5}{*}{\begin{tabular}[c]{@{}l@{}}\textsc{MIMIC}\\ \textsc{-CXR}\end{tabular}} & \textsc{ST}$^{\sharp}$                                        & 0.299          & 0.184          & 0.121          & 0.084          & 0.124          & 0.263          & 0.249             & 0.203            & 0.204            \\
                                                                                       & \textsc{Att2in}$^{\sharp}$                                    & 0.325          & 0.203          & 0.136          & 0.096          & 0.134          & 0.276          & 0.322             & 0.239            & 0.249            \\
                                                                                       & \textsc{AdaAtt}$^{\sharp}$                                    & 0.299          & 0.185          & 0.124          & 0.088          & 0.118          & 0.266          & 0.268             & 0.186            & 0.181            \\
                                                                                       & \textsc{Topdown}$^{\sharp}$                                   & 0.317          & 0.195          & 0.130          & 0.092          & 0.128          & 0.267          & 0.320             & 0.231            & 0.238            \\ \cmidrule(l){2-11} 
                                                                                       & \textsc{Ours}                                                 & \textbf{0.353} & \textbf{0.218} & \textbf{0.145} & \textbf{0.103} & \textbf{0.142} & \textbf{0.277} & \textbf{0.333}    & \textbf{0.273}   & \textbf{0.276}   \\ \bottomrule
\end{tabular}
\vskip -0.3em
\caption{Comparisons of our full model with previous studies on the test sets of \textsc{IU X-Ray} and \textsc{MIMIC-CXR} with respect to NLG and CE metrics.
$\natural$ refers to that the result is directed cited from the original paper and $\sharp$ represents our replicated results by their codes.}
\label{table:nlg-metrics}
\vskip -1em
\end{table*}

\subsection{Baseline and Evaluation Metrics}
\label{subsec:baselines-and-metrics}
To compare with our proposed model, the following ones are used as the main baselines:
\begin{itemize}
    \setlength{\topsep}{0pt}
    \setlength{\itemsep}{0pt}
    \setlength{\parsep}{0pt}
    \setlength{\parskip}{0pt}
    \item \textbf{\textsc{Base}}: this is the vanilla Transformer, with three layers, 8 heads and 512 hidden units without other extensions and modifications.
    \item \textbf{\textsc{Base+rm}}: this is a simple alternative of our proposed model where the relational memory is directly concatenated to the output of the Transformer ahead of the softmax at each time step. This baseline aims to demonstrate the effect of using memory as an extra component instead of integration within the Transformer.
\end{itemize}
In addition, we also compare our model with those in previous studies, including conventional image captioning models, e.g., \textbf{\textsc{ST}} \cite{showandtell}, \textbf{\textsc{Att2in}} \cite{scst}, \textbf{\textsc{AdaAtt}} \cite{adaatt},  \textbf{\textsc{Topdown}} \cite{updown}, and the ones proposed for the medical domain, e.g., \textbf{\textsc{CoAtt}} \cite{coatt}, \textbf{\textsc{Hrgr}} \cite{hrgr} and \textbf{\textsc{Cmas-Rl}} \cite{cmas}.

The performance of the aforementioned models is evaluated by conventional natural language generation (NLG) metrics and clinical efficacy (CE) metrics\footnote{Note that CE metrics only apply to \textsc{MIMIC-CXR} because the labeling schema of CheXpert is designed for \textsc{MIMIC-CXR}, which is different from that of \textsc{IU X-Ray}.}.
The NLG metrics\footnote{\url{https://github.com/tylin/coco-caption}} include
BLEU \cite{bleu}, METEOR \cite{meteor} and ROUGE-L \cite{rouge}.
For clinical efficacy metrics,
we use
the CheXpert \cite{chexpert}\footnote{\url{https://github.com/MIT-LCP/mimic-cxr/tree/master/txt/chexpert}} to label the generated reports and compare the results with ground truths in 14 different categories related to thoracic diseases and support devices.
Precision, recall and F1 are used to evaluate model performance for these metrics.

\subsection{Implementation Details}
\label{subsec:implementation-details}
We adopt the ResNet101 \cite{resnet} pre-trained on Imagenet \cite{imagenet} as the visual extractor to extract patch features with the dimension of each feature set to 2,048.
Note that for \textsc{IU X-Ray}, we use two images of a patient as input to ensure consistency with the experiment settings of previous work.
The Transformer in our proposed model and all baselines are randomly initialized.
For relational memory, its dimension and the number of heads in multi-head attention are set to 512 and 8, respectively, and the number of memory slots is set to 3 by default.
For MCLN, we use two MLPs to obtain $\Delta \boldsymbol{\gamma}$ and $\Delta \boldsymbol{\beta}$ where they do not share parameters.
The model is trained under cross entropy loss with ADAM optimizer \cite{adam}.
We set the learning rate to 5e-5 and 1e-4 for the visual extractor and other parameters, respectively.
We decay such rate by a factor of 0.8 per epoch for each dataset and
set the beam size to 3 to balance the generation effectiveness and efficiency.
Note that the aforementioned hyper-parameters are obtained by evaluating the models on the validation sets of the two datasets.

\section{Results and Analyses}
\label{sec:results-and-analysis}

\subsection{Effect of Relational Memory}
To illustrate the effectiveness of our proposed method,
we experiment with the aforementioned baselines on the two benchmark datasets.
The results are reported in Table \ref{table:baselines}, with \textsc{Base+rm+ mcln} representing our full model (same below).

There are several observations.
First, on NLG metrics, both \textsc{Base+rm} and \textsc{Base+rm+mcln} outperform the vanilla Transformer (\textsc{Base}) on both datasets, 
which confirms the validity of incorporating memory into the decoding process in Transformer because that highly-patternized text in radiology reports are reasonably modeled to some extent.
Second, our full model
achieves the best performance over all baselines on different metrics,
and it particularly outperforms \textsc{Base+rm} with significant improvement,
which clearly indicates the usefulness of MCLN in incorporating memory rather than other ways of integration.
Third, on NLG metrics,
when comparing between the datasets, the performance gains from two memory-driven models (i.e., \textsc{Base+rm} and \textsc{Base+rm+mcln}) over \textsc{Base} on \textsc{IU X-Ray} are larger than that of \textsc{MIMIC-CXR}.
The reason behind might be that
the \textsc{IU X-Ray} is relatively small and patterns among different reports in this dataset are more consistent so that our model helps more with the proposed memory.
Fourthly, on the CE metrics on \textsc{MIMIC-CXR},
our full model shows the same trend as that for NLG metrics,
where it
outperforms all its baselines in terms of precision, recall and F1.
This observation is important because higher NLG scores do not always result in higher clinical scores (e.g., the precision of \textsc{Base+rm} on CE is lower than that of \textsc{Base}),
so that the performance from CE further confirms the effectiveness of our method, whereas compared to \textsc{Base+rm}, MCLN is able to leverage memory in a rather fine-grained way and thus better produce reasonable descriptions for clinical abnormalities.

\subsection{Comparison with Previous Studies}
We compare our full model (denoted as \textsc{Ours}) with existing models on the same datasets, with all results reported in Table  \ref{table:nlg-metrics} on both NLG and CE metrics.
There are several observations drawn from different aspects.
First, Transformer confirms its superiority to sequence-to-sequence structures in this task,
which is illustrated by
the comparison between our models (all baselines and our full model) and \textsc{ST}.
Our full model also outperforms conventional image captioning models, e.g., \textsc{Att2in}, \textsc{AdaAtt} and \textsc{Topdown}, which are designed to generate a short piece of text for an image.
This observation confirms that designing a specific model for long report generation is necessary for this task.
Second, memory shows its effectiveness in this task when 
compared with those complicated models, e.g., \textsc{Hrgr}
uses manually extracted templates.
Particularly, although on the two datasets, reinforcement learning (\textsc{Cmas-RL}) is proved to be the best solution with a careful design of adaptive rewards,
our model achieves the same goal with a simpler method.
Third, 
It is noticed that there are studies, e.g., \textsc{Hrgr},
requires to utilize extra information for this task and
our full model outperforms them without such requirements.
This observation indicates that an appropriate end-to-end design (such as RM and MCLN) of using memory in Transformer 
can alleviate the need for extra resources to enhance this task.

\subsection{Analysis}
We analyze several aspects of our model regarding its hyper-parameters and generation results.
\begin{table}[t]
\centering
\setlength{\tabcolsep}{1.8mm}{\begin{tabular}{@{}c|c|cccc@{}}
\toprule
$\boldsymbol{|\mathcal{S}|}$ & \textsc{\textbf{Para.}} & \textsc{\textbf{BL-1}}  & \textsc{\textbf{BL-2}}  & \textsc{\textbf{MTR}}   & \textsc{\textbf{RG-L}}  \\ \midrule
1             & 76.6M            & 0.350          & 0.217          & 0.141          & 0.278          \\
2             & 81.4M            & 0.355          & 0.215          & 0.141          & 0.278          \\
3             & 86.1M            & \textbf{0.360} & \textbf{0.223} & \textbf{0.144} & 0.279          \\
4             & 90.8M            & 0.354          & 0.217          & 0.142          & \textbf{0.280} \\ \bottomrule
\end{tabular}}
\vskip -0.25em
\caption{
NLG scores of our full model on the \textsc{MIMIC-CXR} test set when different memory slots are used. \textsc{Para.} denotes the number of parameters.}
\label{table:memory-size}
\vskip -1em
\end{table}
\paragraph{Memory Size}
To show the impacts of the memory size, we train RM with different numbers of memory slots, i.e., $|\mathcal{S}| \in\{1, 2, 3, 4\}$ and the results on \textsc{MIMIC-CXR} are shown in Table \ref{table:memory-size}.
In general, since memory size controls how much information is preserved in the past generation steps, it is confirmed in the observation that enlarging memory size by the number of slots results in better overall performance, with 
$|\mathcal{S}|=3$ achieving the best results.
Still, we notice that the overall performance drops when $|\mathcal{S}|=4$, which indicates that too large memory may
introduce redundant and invalid information so as to negatively affect the generation process.
Although enlarging memory size results in increasing parameter numbers, it is demonstrated that there are not too many parameters (comparing to the total number of parameters) introduced whenever adding one slot in the memory.
This observation suggests that the proposed model is effective and efficient in learning with memory for the radiology report generation task.

\paragraph{Report Length}
\begin{figure}[t]
\centering
\includegraphics[width=0.48\textwidth, trim=0 25 0 10]{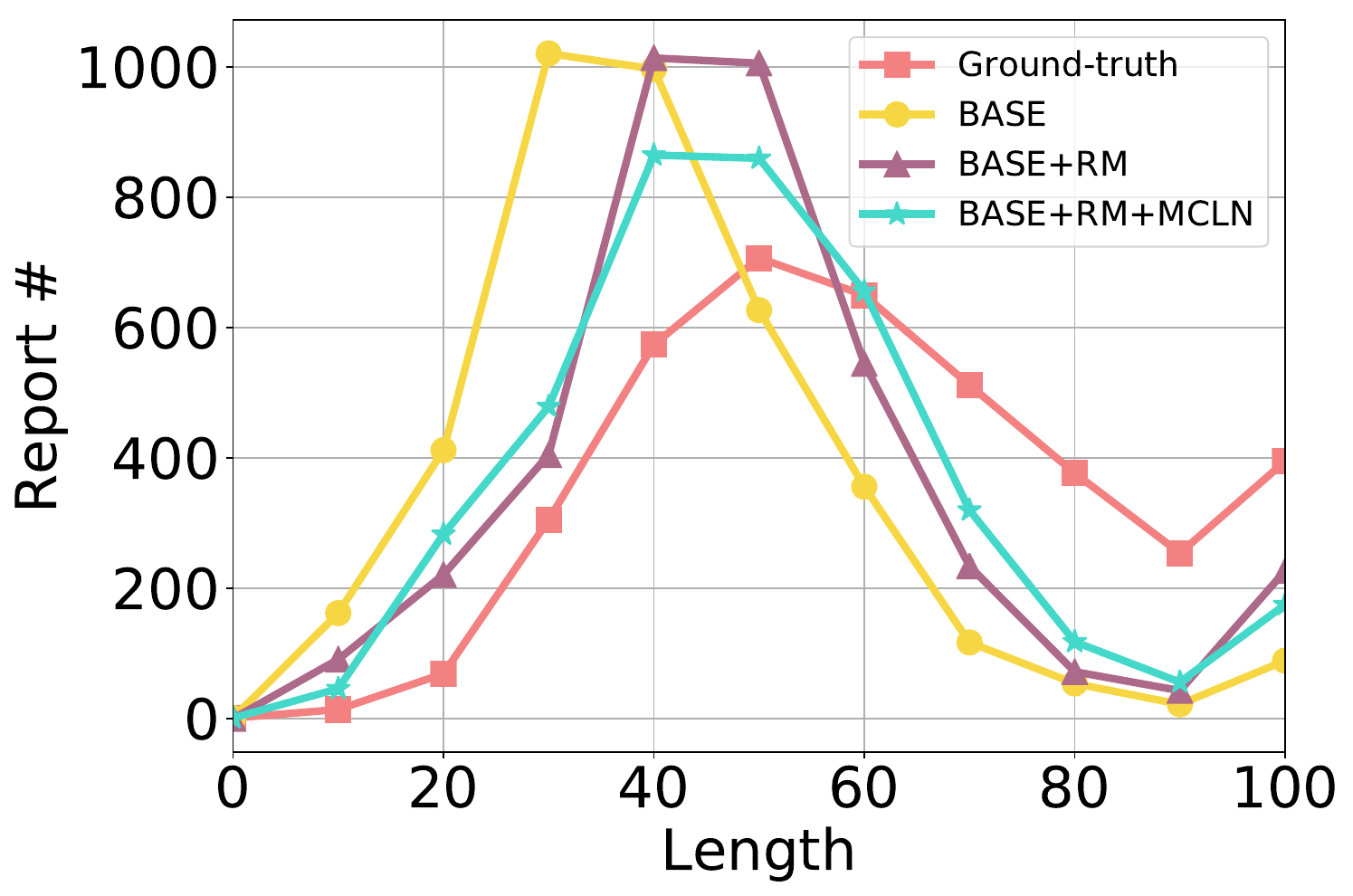}
\caption{The length distributions of the generated reports on the \textsc{MIMIC-CXR} test set from \textsc{Base}, \textsc{Base+ rm} and \textsc{Base+rm+mcln}, as well as the ground-truth.}
\label{fig:length}
\vskip -1em
\end{figure}
\begin{figure*}[t]
\centering
\includegraphics[width=1\textwidth, trim=0 25 0 0]{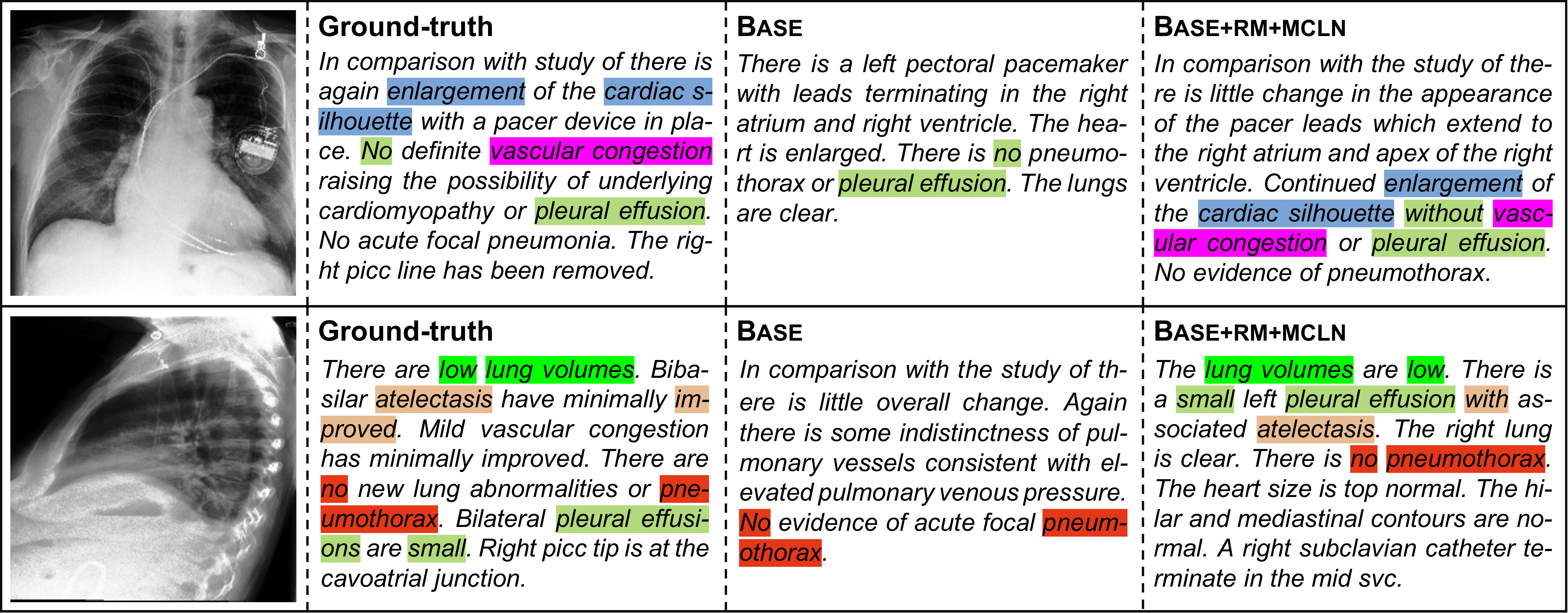}
\caption{Illustrations of reports from ground-truth, \textsc{Base} and \textsc{Base+rm+mcln} models for two X-ray chest images.
To better distinguish the content in the reports, different colors highlight different medical terms.}
\label{fig:qualitative}
\vskip -1.0em
\end{figure*}
In addition to NLG and CE metrics,
another important criterion to evaluate generation models is the length of generated reports comparing to the ground-truth.
In doing so, we categorize all reports generated on the \textsc{MIMIC-CXR} test set into 10 groups (within [0, 100] with interval of 10) according to their round-down lengths and draw curves for their numbers in each category for \textsc{Base}, \textsc{Base+rm} and \textsc{Base+rm+mcln}, as well as the ground-truth.
The results are presented in Figure \ref{fig:length}.
Overall, more reports generated from \textsc{Base+rm} and \textsc{Base+rm+mcln} are longer than that from \textsc{Base} and 
their length distributions are closer to the ground-truth reports,
which thus leads to better evaluation results on NLG metrics.
The reason behind might be that the memory provides more detailed information for the generation process so that the decoder tends to produce more diversified outputs than the original Transformer.
Particularly, 
when comparing \textsc{Base+rm+mcln} and \textsc{Base+rm}, the length distribution of the former generated reports is closer to the ground-truth,
which can be explained by that,
instead of applying memory to the final output, leveraging memory at each layer in Transformer is more helpful and thus controls the decoding process in a fine-grained way.
The above observations show that both memory and the way of using it are two important factors to enhance radiology report generation.

\vspace{-0.2cm}
\paragraph{Case Study}
To further investigate the effectiveness of our model, we perform qualitative analysis on some cases with their ground-truth and generated reports from different models.
Figure \ref{fig:qualitative} shows two examples of front and lateral chest X-ray images from \textsc{MIMIC-CXR} and such reports,
where different colors on the texts indicate different medical terms.
It is observed in these cases that \textsc{Base+rm+mcln} is able to generate descriptions aligned with that written by radiologists with similar content flow.
For example,
in both cases, patterns in the generated reports follow the structure that
starting from reporting abnormal findings (e.g., ``\textit{cardiac silhouette}'' and ``\textit{lung volumes}''), and then concluding with potential diseases (e.g., ``\textit{pleural effusion}'' and ``\textit{atelectasis}'').
In addition,
for the necessary medical terms in the ground-truth reports,
\textsc{Base+rm+mcln} covers almost all of them in its generated reports while
vanilla Transformer did much worse,
e.g., the key terms ``\textit{enlarged cardiac silhouette}'', ``\textit{atelectasis}'' and ``\textit{small pleural effusion}'' in the two examples are not generated.
\begin{figure*}[t]
\centering
\includegraphics[width=1\textwidth, trim=0 25 0 0]{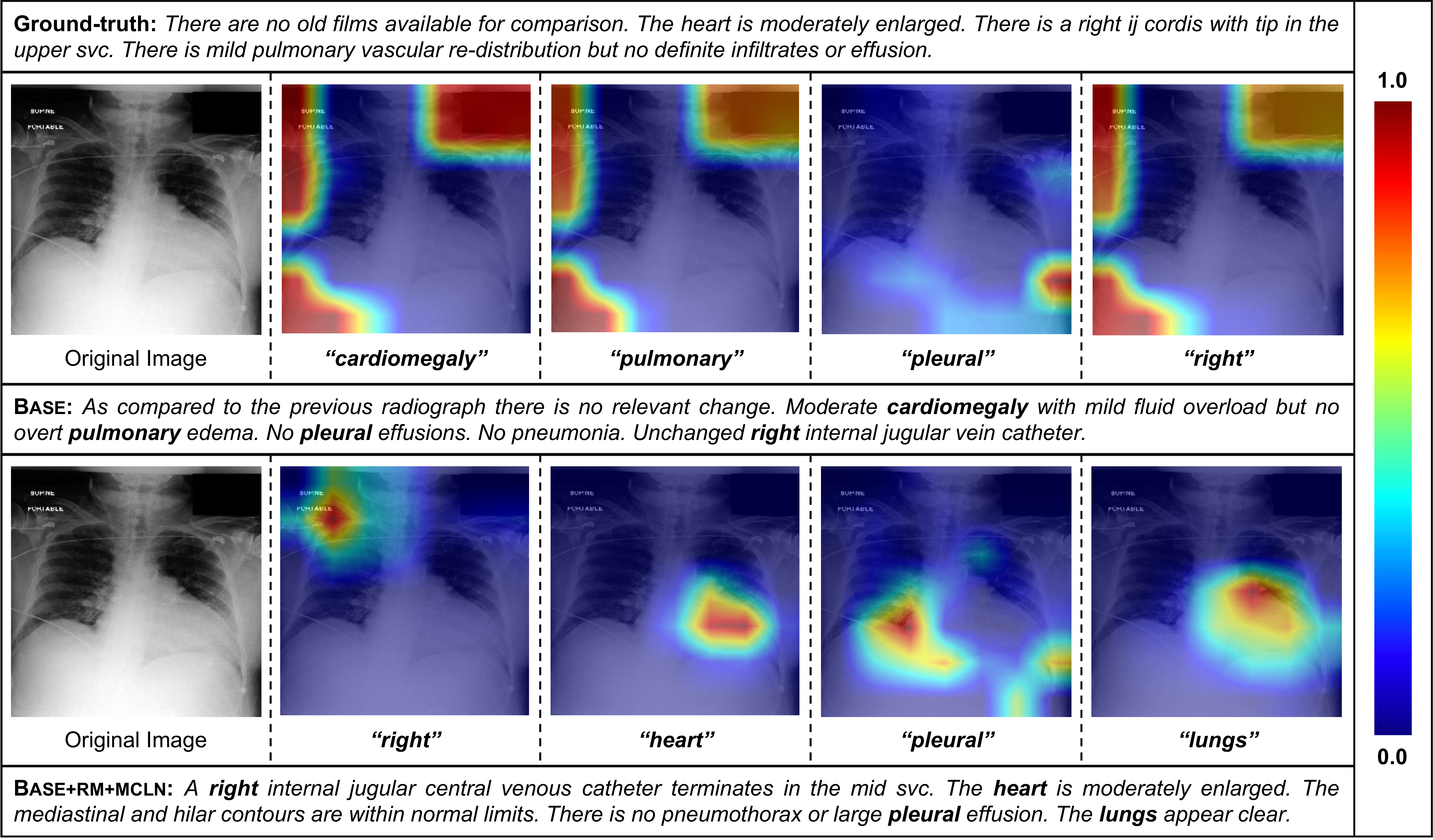}
\caption{Visualizations of image-text attention mappings between a specific chest X-ray and generated reports from \textsc{Base} and \textsc{Base+rm+mcln}, respectively.
Colors from blue to red represent the weights from low to high.}
\label{fig:attention}
\vskip -1em
\end{figure*}

To further investigate different models qualitatively,
we randomly select a chest X-ray on the \textsc{MIMIC-CXR} test set and visualize the image-text attention mappings from \textsc{Base} and \textsc{Base+rm+mcln}.
Figure \ref{fig:attention} shows the intermediate image-text correspondences for several words from the multi-head attentions in the first layer of the decoders.
It is observed that \textsc{Base+rm+mcln} is better at aligning the locations with the indicated disease or parts.
This observation suggests that our model not only enhances the power of radiology report generation, but also improves the interaction between the images and the generated texts.

\vspace{-0.2cm}
\paragraph{Error Analysis}
To analyze the errors from our model, especially in targeting the low CE scores,
it is found that the class imbalance is severe on the datasets and affects the model training and inference, where majority voting is observed in the generation process.
For example, on MIMIC-CXR, consolidation only accounts for 3.9\% in the training set so that the trained model only recognizes that 2.9\% results in this case compared with the ground truth 6.3\%.
Thus how to address the data bias problem is a possible future work to improve the accuracy of the generated radiology reports.

\vspace{-0.1cm}
\section{Related Work}
\label{sec:relatedwork}
\vspace{-0.1cm}
The most popular related task to ours is image captioning \cite{showandtell,xu2015show,updown,wang2019hierarchical}, which aims to describe images with sentences.
Different from them, radiology report generation requires much longer generated outputs,
and possesses other features such as patterns, so that this task has its own characteristics requiring particular solutions.
For example,
\newcite{coatt} proposed a co-attention mechanism and leveraged a hierarchical LSTM to generate reports.
\newcite{hrgr,kerp}
proposed to use a manually extracted template database to help generation with bunches of special techniques to utilize templates.
\newcite{clinically} proposed an approach with reinforcement learning to maintain the clinical accuracy of generated reports.
Compared to these studies, our model offers an alternative solution to this task with an effective and efficient enhancement of Transformer via memory.

Extra knowledge (e.g., pre-trained embeddings 
\cite{song2017learning,song2018complementary,zhang2019multiplex} and pretrained models \cite{devlin2019bert,diao2019zen}) can provide useful information and thus enhance model performance for many NLP tasks \cite{tian2020joint,tian-etal-2020-suppertagging,tian-etal-2020-constituency}.
Specifically, memory and memory-augmented neural networks \cite{zeng2018topic,relational,diao2020keyphrase,tian2020improving} are another line of related research, which can be traced back to
\newcite{memory}, which proposed memory networks to leverage extra information for question answering;
then \newcite{end2end} improved it with an end-to-end design to ensure the model being trained with less supervision.
Particularly for Transformer, there are also memory-based methods proposed.
For example,
\newcite{productkeys} proposed to solve the under-fitting problem of Transformer by introducing a product-key layer that is similar to a memory module.
\newcite{memo} proposed MEMO, an adaptive memory to reason over long-distance texts.
Compared to these studies,
the approach proposed in this paper focuses on leveraging memory for decoding rather than encoding,
and presents a relational memory 
to learn from previous generation processes as well as patterns for long text generation.
To the best of our knowledge, this is the first study incorporating memory for decoding with Transformer and applied for a particular task,
which may provide a reference for studies in the line of this research.

\vspace{-0.2cm}
\section{Conclusion}
\label{sec:conclusion}
\vspace{-0.2cm}
In this paper, we propose to generate radiology reports with
memory-driven Transformer,
where a relational memory is used to record the information from previous generation processes and a novel layer normalization mechanism is designed to incorporate the memory into Transformer.
Experimental results on two benchmark datasets illustrate the effectiveness of the memory by either concatenating it with the output or integrating it with different layers of the decoder by MCLN, which obtains the
state-of-the-art performance.
Further analyses investigate how memory size affects model performance
and show that our model is able to generate long reports with necessary medical terms and meaningful image-text attention mappings.

\bibliography{anthology,emnlp2020}
\bibliographystyle{acl_natbib}

\end{document}